\documentclass{article}

\PassOptionsToPackage{numbers, compress}{natbib}

\usepackage[preprint]{neurips_2020}

\usepackage[utf8]{inputenc} 
\usepackage[T1]{fontenc}    
\usepackage{hyperref}       
\usepackage{url}            
\usepackage{booktabs}       
\usepackage{amsfonts}       
\usepackage{nicefrac}       
\usepackage{microtype}      
\usepackage[pdftex]{graphicx}
\usepackage{amsmath}
\title{Meta Graph Attention on Heterogeneous Graph with Node-Edge Co-evolution}

\author{
	Yucheng Lin$^*$, ~Huiting Hong \thanks{Both authors contributed equally to this research.}, ~Xiaoqing Yang, ~Xiaodi Yang, ~Pinghua Gong, ~Jieping Ye\\
	Didi Chuxing \\
	\texttt{\{linyucheng, honghuiting, yangxiaoqing, yangxiaodi\_i,}\\
	\texttt{gongpinghua, yejieping\}@didiglobal.com}\\
}

\begin{document}

\maketitle

\begin{abstract}
  Graph neural networks have become an important tool for modeling structured data.  In many real-world systems, intricate hidden information may exist, e.g., heterogeneity in nodes/edges, static node/edge attributes, and spatiotemporal node/edge features. However, most existing methods only take part of the information into consideration. In this paper, we present the Co-evolved Meta Graph Neural Network (CoMGNN), which applies meta graph attention to heterogeneous graphs with co-evolution of node and edge states. We further propose a spatiotemporal adaption of CoMGNN (ST-CoMGNN) for modeling spatiotemporal patterns on nodes and edges. We conduct experiments on two large-scale real-world datasets. Experimental results show that our models significantly outperform the state-of-the-art methods, demonstrating the effectiveness of encoding diverse information from different aspects.
\end{abstract}

\section{Introduction}
The graph structure models a set of objects (nodes) and their relationships (edges), which shows strong expressiveness across various fields. Analyzing graphs and mining the latent information in structured data is of great importance in many real-world tasks. 

In the past few years, graph neural networks (GNNs) \cite{survey1} have received increasing attentions across various areas including social network \cite{GraphSage}, recommendation system \cite{hetsann}, knowledge graph \cite{kg}, traffic forecasting\cite{DCRNN,STGCN}, etc. Most existing GNN methods focus on certain types of graphs. Early methods \cite{ChebNet,GCN,GraphSage,GAT} are usually based on simple homogeneous graphs, which contain only one node type and one edge type, e.g., citation networks \cite{Cora}(``paper'' as node, ``cite'' as edge) and social networks (``human'' as node, ``friend to'' as edge). Moreover, only static node features are provided as input in these works. Recent works pay attention to some harder cases and include more useful information as input. 

However, when applying GNNs to solve real-world problems, we often encounter complicated issues that existing methods could not handle them all:

\begin{itemize}
\item \textbf{Heterogeneity in graph.} A real-world system often involves multi-type of objects or relations. Therefore, it should be represented by a heterogeneous graph. However, learning the distinct properties between various types of nodes or edges is challenging.

\item \textbf{Consider node and edges features simultaneously.} With node features as input, most existing GNN methods focus on node prediction problems by aggregating information from neighboring nodes layer by layer. However, features on edges are usually neglected, which does not take full advantage of rich information indicating detailed patterns of systems. Moreover, prediction tasks on edges should be considered.

\item \textbf{Input features vary over time.} Spatiotemporal data prediction by GNN is also an important topic, where some node (or edge) features could change over time. For these problems, we should take both spatial patterns and temporal dependencies into account.

\item \textbf{Scalability on large-scale data.} In applications of industrial systems, size of graphs and capacity of features are usually quite large. Thus, the efficiency and scalability of algorithms on large-scale data should be taken into account.
\end{itemize}

Generally speaking, a real-world system may involve one or more challenges as listed above. For example, in order to represent a classic recommendation system \cite{MovieLens}, we should at least build a large-scale bipartite graph (a special heterogeneous graph). And for a traffic forecasting problem, we should construct a large-scale urban road graph, where each node represents a road segment and each edge indicates adjacency between two road segments. Both static and dynamic features exist on nodes and edges. For example, on nodes, there are  static properties like length or width of road segments and spatiotemporal features like real-time traffic status. In addition, features on the edge include static attributes such as turning directions (like straight forward, turn left) and spatiotemporal information like real-time passing flow.


In this paper, we propose a general framework entitled CoMGNN to address the aforementioned challenges. The contributions of this work are summarized as follows:
\begin{itemize}
  \item We propose a novel deep graph neural network based model named CoMGNN, which can utilize both node and edge attributes on large-scale heterogeneous graphs. In addition, both downstream node and edge prediction tasks are supported.
  \item Meta learning and heterogeneous attention mechanism are applied to better aggregate heterogeneous information through different types of edges. 
  \item We further propose an enhanced version of CoMGNN for spatiotemporal prediction called ST-CoMGNN. To the best of our knowledge, this is the first to model heterogeneous graphs with spatiotemporal features on both nodes and edges.
  \item We conduct extensive experiments on two large-scale real-world datasets collected from a ride-hailing platform and the experimental results demonstrate the effectiveness of CoMGNN and ST-CoMGNN.
\end{itemize}






\section{Related Work}
Early graph convolution methods, based on the graph spectral theory \cite{spectal}, define convolution operations in the Fourier domain (or spectral domain). Spectral models like ChebNet \cite{ChebNet} and GCN \cite{GCN} rely on Laplacian matrices to analyze graphs, which requires the adjacency matrices to be symmetric. Therefore, they are only suitable for undirected graphs. Then, non-spectral methods \cite{survey1} like GraphSage \cite{GraphSage} and GAT \cite{GAT} are proposed, which do not depend on the calculation of Laplacian matrices and show an improvement of performance on directed graphs. These classical methods focus on homogeneous graphs and update the hidden states of nodes layer by layer, without any consideration of edge information.

In the past several years, researches began to explore more aspects of graph information. On the one hand, EGNN \cite{EGNN} exploits edge features by considering multi-dimensional edge features as multi-channel signals and applies a graph attention operation to each channel separately. \cite{DGT} formulates a multi-attributed graph translation problem and proposes a multi-block translation architecture called NEC-DGT to tackle this problem, in which the hidden edge and node states are co-evolved.
On the other hand, methods like R-GCN\cite{rgcn}, HAN\cite{han}, HetSANN\cite{hetsann} have been proposed to analyze the heterogeneity in nodes or edges. However, none of them takes both heterogeneity and node-edge co-evolution into consideration.

As for the problem of spatiotemporal prediction, early methods such as DCRNN\cite{DCRNN}, STGCN\cite{STGCN}, Graph WaveNet\cite{GraphWaveNet} only deal with node features on homogeneous graph. Lately, ASTGCN\cite{ASTGCN} considers heterogeneity in the time domain and ST-MetaNet\cite{st-metanet} encodes both edge and node attributes by applying meta-learning based graph attention to homogeneous graphs.

All these works fail to encode some important information on the graph, which may be quite helpful for target tasks.  


\section{Methodology}
\begin{figure}
\centering
\includegraphics[width=\textwidth]{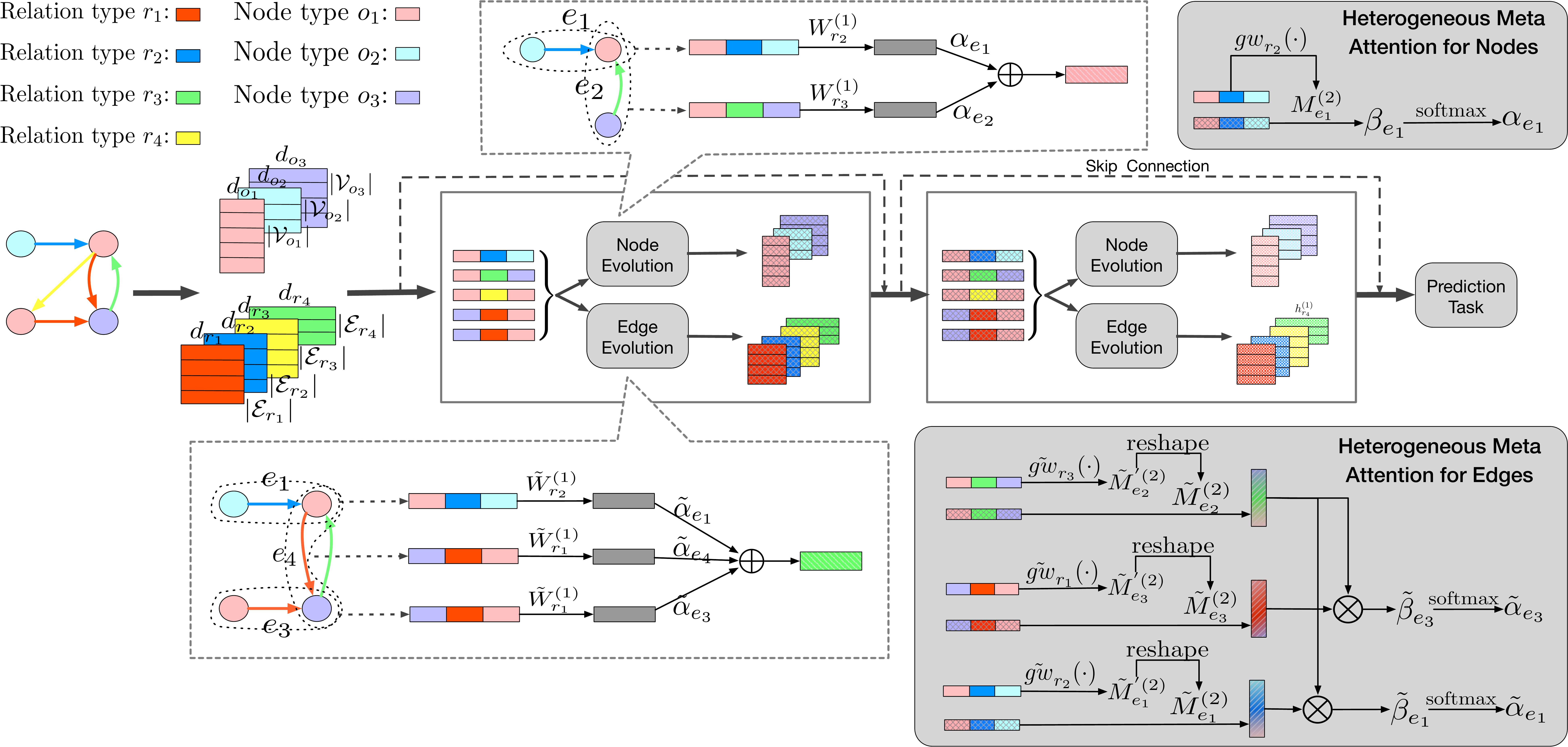}
\caption{The Vanilla Architecture of Co-evolution Meta Graph Neural Network (CoMGNN).}
\label{fig:base_model}
\end{figure}
\subsection{Problem Definition}
A graph/network comprises a set of vertices/nodes $\mathcal{V}$ and a set of edges $\mathcal{E}$ between vertices, denoted as $\mathcal{G}=(\mathcal{V}, \mathcal{E})$. When there are multiple types of nodes or edges/relations in the graph, we call the graph as heterogeneous and denote the set of node and edge types by $\mathcal{O}$ and $\mathcal{R}$, respectively. Each vertex $v\in\mathcal{V}$ is mapped to a certain type $o\in{\mathcal{O}}$ by the node type mapping function $o=\phi(v)$. Each edge $e\in\mathcal{E}$ is associated with two vertices and a relation type, i.e., $e=(v_i,v_j,r)$, where $r\in\mathcal{R}$ is obtained by the edge type mapping function $r=\psi(e)$. It is common to refer to the graph with node attributes as an attributed network, i.e., $G=(\mathcal{V}, \mathcal{E}, \mathcal{A}_{\mathcal{V}})$. In this paper, we name the graph with both node attributes and edge attributes as a multi-attributed network, denoted as $G=(\mathcal{V}, \mathcal{E}, \mathcal{A}_{\mathcal{V}}, \mathcal{A}_{\mathcal{E}})$. Note that different types of nodes can have different attribute sets, and so do the attributes of edges. We denote the attribute set of the node type $o_i$ by $\mathcal{A}_{o_i}\in \mathbb{R}^{|\mathcal{V}_{o_i}|\times d_{o_i}}$, where $|\mathcal{V}_{o_i}|$ is the number of vertices belonging to the node type $o_i$ and $d_{o_i}$ is the dimensionality of attributes of the node type $o_i$. Analogously, $\mathcal{A}_{r_i}\in \mathbb{R}^{|\mathcal{E}_{r_i}|\times d_{r_i}}$ denotes the attributes of edges belonging to relation type $r_i$, where $d_{r_i}$ is the dimensionality of attributes of the relation type $r_i$. In this paper, our aim is to embed the multi-attributed heterogeneous graph to boost specific task performance.

\subsection{Proposed method: CoMGNN}
Different with previous GNN models which focus on aggregating node information from neighboring vertices, we propose a Co-evolution Meta Graph Neural Network (CoMGNN) to aggregate node and edge information together. It is important for some prediction tasks to learn meaningful representations of edges, such as recommendation task. A common way is to learn the representation of vertices firstly, then concatenate representations of two vertices of each edge as the edge representation. However, it does little to track edge attributes in a graph. To overcome the limitations of node aggregation, our CoMGNN co-evolves the hidden states of nodes and edges. As shown in Figure \ref{fig:base_model}, CoMGNN stacks aggregation layers, which consists of two evolution functions to update the hidden states of nodes and edges, respectively.
\paragraph{Node evolution.}
The evolution function of nodes is to aggregate information of neighboring nodes and edges as follows:
\begin{equation}
\hat{h}_{v_i}^{(l)}=\sum_{e_k=(v_i, v_j, \psi(e_k))\in{\mathcal{E}_{v_i}}} \alpha_{e_k}*f(h^{(l-1)}_{v_i}, h^{(l-1)}_{v_j}, h^{(l-1)}_{e_k})
\label{fun:node_aggregation}
\end{equation}
where $\mathcal{E}_{v_i}$ denotes the set of edges linking to vertex $v_i$, and $h^{(l-1)}_{v_i}\in\mathbb{R}^{d^{(l-1)}_{\phi(v_i)}}$ means the hidden state of vertex $v_i$ in $(l\!-\!1)$-th layer. In the $0$-th layer, node state is initialized by its input attributes, i.e., $h^{(0)}_{v}=a_v$, where $a_v \in \mathcal{A}_{\phi({v})}$. Similarly, the $0$-th layer's edge state is $h^{(0)}_{e}=a_{e}\in \mathcal{A}_{\psi(e)}$. We utile attention mechanism to compute weights of edges $\alpha_{e_k}$ to the target vertex $v_i$, which is elaborated in Section \ref{sec:meta_attention} below.
The mapping function $f$ maps the information of edges, which have different sizes of dimensionality, into the same low-dimensional vector space according to their relation types:
\begin{equation}
f(h^{(l-1)}_{v_i}, h^{(l-1)}_{v_j}, h^{(l-1)}_{e_k})= W^{(l)}_{\psi(e_k)}\left[h^{(l-1)}_{v_i}\| h^{(l-1)}_{e_k}\|h^{(l-1)}_{v_j}\right]+ b^{(l)}_{\psi(e_k)}
\end{equation}
where $\|$ indicates the concatenate operation. $W^{(l)}_{\psi(e_k)}\in \mathbb{R}^{\hat{d}^{(l)}_{\mathcal{V}} \times \hat{d}^{(l-1)}_{\psi(e_k)}}$ is a parameter matrix of $l$-th layer associated with the relation type $r\!=\!\psi(e_k)$, where $\hat{d}^{(l-1)}_{\psi(e_k)}=d^{(l-1)}_{\phi(v_i)}+d^{(l-1)}_{\psi(e_k)}+d^{(l-1)}_{\phi(v_j)}$. And $\hat{d}^{(l)}_{\mathcal{V}}$ is the dimensionality of common low-dimensional vector space for node evolution in $l$-th layer. After aggregating the neighborhood information, we consider the information from the vertex itself by skip connection:
\begin{equation}
h^{(l)}_{v_i} = \text{Mish}\left(W^{(l)}_{\phi(v_i)}\left[\hat{h}_{v_i}^{(l)}\|h^{(l-1)}_{v_i}\right] + b^{(l)}_{\phi(v_i)}\right)
\end{equation}
where $W^{(l)}_{\phi(v_i)}\in \mathbb{R}^{d^{(l)}_{\phi(v_i)}\times\left(\hat{d}^{(l)}_{\mathcal{V}}+d^{(l-1)}_{\phi(v_i)}\right)}$, and $b^{(l)}_{\phi(v_i)}\in \mathbb{R}^{d^{(l)}_{\phi{(v_i)}}}$ are trainable parameters. Note that, Mish \cite{mish} is applied as the non-monotonic activation function to help with effective optimization.

\paragraph{Edge evolution.} The evolution function of edges $e_k=(v_i, v_j,\psi(e_k))$ ($e_k=(v_i,v_j)$, for short) is to aggregate the neighborhood information of the edge's two related vertices:
\begin{equation}
\hat{h}^{(l)}_{e_k} = \sum_{\underset{(v_i, v_q)\in \mathcal{E}_{v_i}}{e_n=}} \tilde{\alpha}_{e_n} * \tilde{f}(h^{(l-1)}_{v_i}, h^{(l-1)}_{v_q}, h^{(l-1)}_{e_n}) + \sum_{\underset{(v_j, v_p)\in \mathcal{E}_{v_j}}{e_m=}} \tilde{\alpha}_{e_m} * \tilde{f}(h^{(l-1)}_{v_j}, h^{(l-1)}_{v_p}, h^{(l-1)}_{e_m})
\label{fun:edge_aggregate}
\end{equation}
where $\tilde{\alpha}_{e}$ is the weight of the neighboring edge $e$ to the target edge $e_k$, computed by the attention mechanism in Section \ref{sec:meta_attention}.
The mapping function $\tilde{f}$ is the same as $f$ of node evolution, but with different sets of parameters $\tilde{W}^{(l)}_{\psi(e)}\in \mathbb{R}^{\hat{d}^{(l)}_{\mathcal{E}}\times\hat{d}^{(l-1)}_{\psi(e)}}, \tilde{b}^{(l)}_{\psi(e)}\in \mathbb{R}^{\hat{d}^{(l)}_{\mathcal{E}}}, \forall{l}\in \{1,\ldots,L\}$. Likewise, the edge representation in $l$-th layer combines the edge information of itself and aggregated information:
\begin{equation}
h^{(l)}_{e_k} = \text{Mish}\left(\tilde{W}^{{(l)}{\star}}_{{\psi(e_k)}} \left[\hat{h}^{(l)}_{e_k}\|h^{(l-1)}_{e_k}\right] + \tilde{b}^{{(l)}\star}_{\psi(e_k)}\right)
\end{equation}
where $\tilde{W}^{{(l)}{\star}}_{{\psi(e_k)}}\in \mathbb{R}^{d^{(l)}_{\psi(e_k)}\times\left(\hat{d}^{(l)}_{\mathcal{E}}+d^{(l-1)}_{\psi(e_k)}\right)}$, and the superscript $\star$ is used to distinguish $\tilde{W}^{{(l)}{\star}}_{{\psi(e_k)}}$ and $\tilde{b}^{{(l)}\star}_{\psi(e_k)}$ from other parameter matrices mentioned ahead.

\subsubsection{Heterogeneous Meta Attention Mechanism}\label{sec:meta_attention}
The correlations between target vertex/edge and its neighborhood information are related to not only the node/edge type, but also the attributes. For example, in a urban road graph, although nodes are of the same type meaning road segments, some latent differences do exist among road segments with distinct widths, free flow speeds, etc.

\paragraph{Node meta attention.}
Given a target vertex $v_i$, the attention score of its $1$-hop neighboring information $(h^{(l-1)}_{v_i}, h^{(l-1)}_{v_j}, h^{(l-1)}_{e_k})$ in Eq. \eqref{fun:node_aggregation} depends on a target-specific vector $M^{(l)}_{e_k}$ and bias $b^{(l)}_{e_k}$:
\begin{equation}
\beta_{e_k} = \text{LeakyReLU}\left({M^{(l)}_{e_k}}^{\top} {\left[h^{(l-1)}_{v_i}\|h^{(l-1)}_{e_k}\|h^{(l-1)}_{v_j}\right]} + b^{(l)}_{e_k}\right).
\end{equation}
$M^{(l)}_{e_k}$ and $b^{(l)}_{e_k}$ indicate the preference of the target $v_i$ to the edge $e_k$, which is encoded by the meta learner:
\begin{equation}
M^{(l)}_{e_k} = gw_{{\psi(e_k)}}\left({MK}_{e_k}\right)\in\mathbb{R}^{\hat{d}^{(l-1)}_{\psi(e_k)}}, \quad b^{(l)}_{e_k}=gb_{{\psi(e_k)}}\left({MK}_{e_k}\right)\in\mathbb{R},
\label{fun:meta-learner}
\end{equation}
where the meta knowledge ${MK}_{e_k}=a_{v_i}\|a_{e_k}\|a_{v_j}$. The meta learner is a composition of functions $gw_{{\psi(e_k)}}(\cdot)$ and $gb_{{\psi(e_k)}}(\cdot)$, which are implemented with fully connected networks.
Then we normalize the attention score $\beta_{e_k}$ over the target $v_i$'s neighborhood information:
\begin{equation}
\alpha_{e_k} = \text{softmax}(\beta_{e_k})= \frac{\exp(\beta_{e_k})}{\sum_{e_n\in\mathcal{E}_{v_i}}\exp(\beta_{e_n})}.
\end{equation}


\paragraph{Edge meta attention.}
Given a target edge $e_k$, the attention score of its $1$-hop neighboring information $(h^{(l-1)}_{v_i}, h^{(l-1)}_{v_q}, h^{(l-1)}_{e_n})$ in Eq. \eqref{fun:edge_aggregate} depends on hidden states of $e_k$ and $e_n$:
$\tilde{\beta}_{e_n} = {\tilde{h}^{(l)\top}_{e_k}}\tilde{h}^{(l)}_{e_n}$,
where $\tilde{h}^{(l)}_{e_k}$ and $\tilde{h}^{(l)}_{e_n}$ are generated by the transformation:
\begin{equation}
\tilde{h}^{(l)}_{e_k} = \sigma\left(\tilde{M}^{(l)}_{e_k}{\left[h^{(l-1)}_{v_i}\|h^{(l-1)}_{e_k}\|h^{(l-1)}_{v_j}\right]}\right),\qquad
\tilde{h}^{(l)}_{e_n} = \sigma\left(\tilde{M}^{(l)}_{e_n}{\left[h^{(l-1)}_{v_i}\|h^{(l-1)}_{e_n}\|h^{(l-1)}_{v_q}\right]}\right),
\end{equation}
where $\sigma$ is the sigmoid function. $\tilde{M}^{(l)}_{e_k}$/$\tilde{M}^{(l)}_{e_n}$ indicates the transformation matrix from meta knowledge $e_k$/$e_n$ to a common vector space for attention, which is encoded by meta learner:
\begin{equation}
\tilde{M}^{(l)}_{e_k}=\tilde{gw}_{\psi(e_k)}\left({MK}_{e_k}\right)\in \mathbb{R}^{d^{(l)}\times \hat{d}^{(l-1)}_{\psi(e_k)}},\qquad
\tilde{M}^{(l)}_{e_n}=\tilde{gw}_{\psi(e_n)}\left({MK}_{e_n}\right)\in \mathbb{R}^{d^{(l)}\times\hat{d}^{(l-1)}_{\psi(e_n)}}.
\end{equation}
The fully connected network $\tilde{gw}_{\psi(e_k)}(\cdot)$/$\tilde{gw}_{\psi(e_n)}(\cdot)$ is shared with the same edge type $\psi(e_k)$/$\psi(e_n)$. Note that $\tilde{M}^{(l)}_{e_k}$ is reshaped from the output of the fully connected network, whose output is a vector.
Then, the weight $\tilde{\alpha}_{e_n}$ in edge evolution (Eq. \eqref{fun:edge_aggregate}) is the normalization of the attention score across the target $e_k=(v_i, v_j)$'s neighborhood information:
\begin{equation}
\tilde{\alpha}_{e_n} = \text{softmax}(\tilde{\beta}_{e_n})= \frac{\exp(\tilde{\beta}_{e_n})}{\sum_{e_m\in\{\mathcal{E}_{v_i},\mathcal{E}_{v_j}\}}\exp(\tilde{\beta}_{e_m})},
\end{equation}


\subsection{Spatiotemporal extension of CoMGNN: ST-CoMGNN}
In a spatiotemporal prediction problem, input features on vertices or edges change over time. In this section, we propose Spatiotemporal Co-evolution Meta Graph Neural Network(ST-CoMGNN), which is an extension of CoMGNN suitable for spatiotemporal prediction problems.

\begin{figure}
\centering
\includegraphics[width=\textwidth]{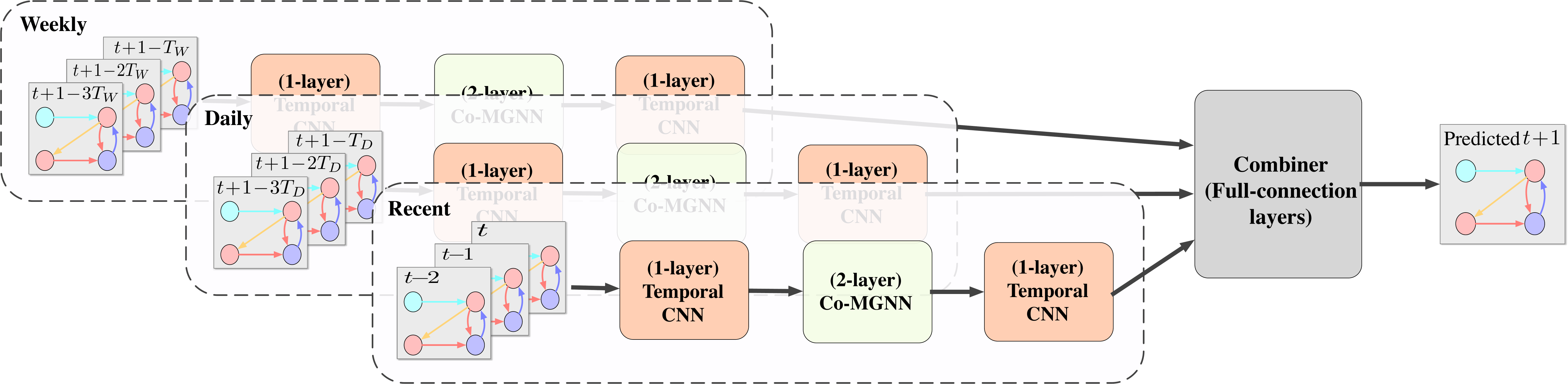}
\caption{An illustration of spatiotemporal CoMGNN (ST-CoMGNN).}
\label{fig:st_model}
\end{figure}

As shown in Figure~\ref{fig:st_model}, inspired by STGCN\cite{STGCN}, each component in ST-CoMGNN is formed as a ``sandwich'' structure with two temporal convolution layers and one spatial layer in the middle. The temporal convolution layer models temporal dynamics on vertices/edges through a 1-D convolution followed by gated linear units (GLUs) as non-linearity. The spatial layer uses $k$ stacked CoMGNNs to catch spatial dependencies in the heterogeneous graph. To model heterogeneity in the time domain, ST-CoMGNN includes three components for time periods in different scales, i.e., recent (5-minute-periodic), daily-periodic and weekly-periodic. These components are of the same structure, where input features are formed in different time intervals, and the output states are combined to make the final prediction.
Note that, in our proposed ST-CoMGNN, both spatiotemporal node and edge features are allowed for input.




\section{Experiments}
In this section, we evaluate our proposed CoMGNN on two tasks: a recommendation task which involves heterogeneous information and multi-attributes, and a traffic prediction task which involves spatio-temporal data. We first introduce the details of recommendation scenario and conduct comparison experiments. Then we evaluate the performance of ST-CoMGNN on traffic prediction task compared to other state-of-the-art methods, followed by ablation tests. The proposed CoMGNN is implemented with Tensorflow toolbox\footnote{We will open source codes of CoMGNN in the camera-ready version.}, and all experiments are conducted on a 64-bit machine with Nvidia Tesla P100 GPU.

\subsection{Hitch Ride Route Matching}
\begin{figure}
\centering
\includegraphics[width=\textwidth]{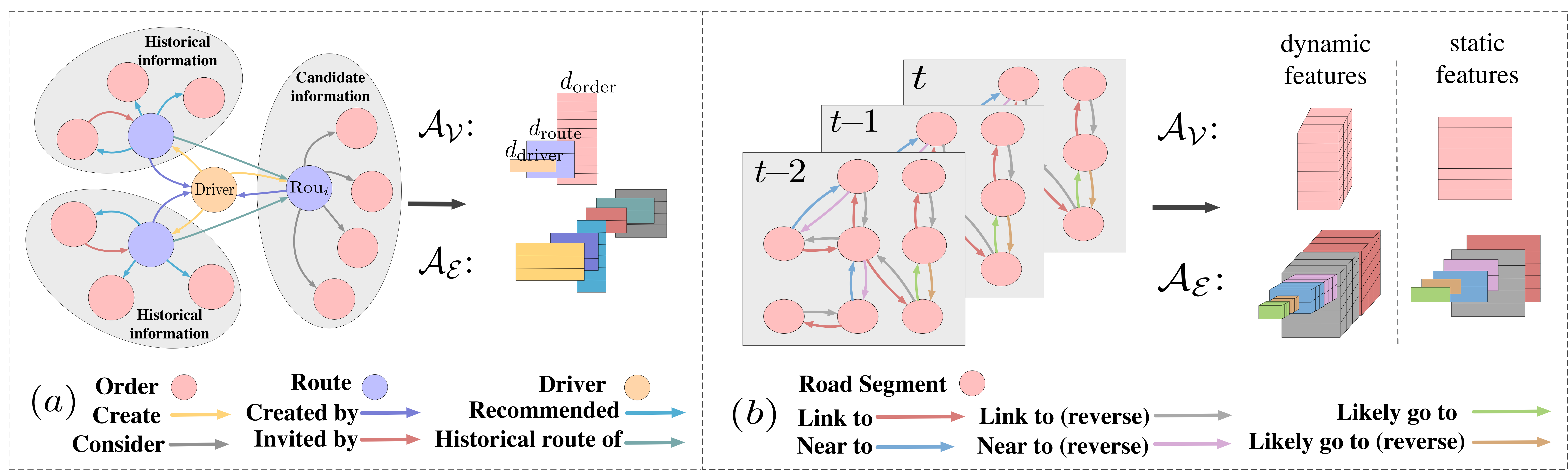}
\caption{The multi-attributed heterogeneous networks on (a) the hitch ride application and (b) the geographical map for traffic prediction.}
\label{fig:hitch_ride}
\end{figure}

On a hitch ride service platform, a driver creates a route that she/he plans to travel and passengers place order requests that they want to go. Then the driver will invite some order from the order list in the platform to drop off the passengers to their destinations and get paid. To improve efficiency of matching between routes and orders, the hitch ride platform needs to provide the driver with a good recommendation order list according to the route requirement. We collect $478,537$ routes and $944,355$ orders from a hitch ride service platform. For each target $\text{route}_i$ to recommend, we construct a multi-attributed heterogeneous network, connecting $\text{route}_i$ with its driver, latest $\mu$ historical routes and orders, which is depicted in the left part of Figure \ref{fig:hitch_ride}. The node attributes of drivers are demographic features, and the attributes of routes and orders are travel requirements, including origin, destination, departure time, etc. The edge attributes of ``create'' and ``be created by'' include day of the week and hour of the route create time. Finally, the edge attributes between route and order are calculated information, such as the detour distance and the difference of their departure times.
Note that a driver may have historical routes, and they can be connected to the target route with similarity information (edge attributes of ``historical route of''), such as the interval between create time of two routes. The relation ``consider'' means that the target route considers the order as a candidate order. Our task is to rank the candidate orders and recommend top-k orders to the target route. And the label is whether the route invited the order or not.

\paragraph{Experimental setup.} We compare against two inductive GNN-based baselines: GraphSage \cite{GraphSage} and NEC-DGT \cite{DGT}. GraphSage considers graphs as the homogeneous graph with node attributes. We provide the same graph with Figure \ref{fig:hitch_ride} for GraphSage, excluding the edge attributes and the type information of nodes and edges. The node attributes of GraphSage are the concatenated vector of attributes of all node types, padding zeros for missing attributes. Mean aggregator and LSTM aggregator are used in GraphSage-mean and GraphSage-LSTM, respectively. And we omit the type information of nodes and edges of the graph in Figure \ref{fig:hitch_ride} to be the input of NEC-DGT, which is designed for multi-attributed homogeneous graph. For all GNN-based models, the number of layers $L\!=\!2$, with $d^{(1)}\!=\!32$ neurons and $d^{(2)}\!=\!32$ respectively. For our $2$-layer CoMGNN, the dimension of all types of nodes $d^{(l)}_{o}$ and edges $d^{(l)}_{r}$ are set to $32$, and the size of common vector space $\hat{d}^{(l)}_{\mathcal{V}}\!=\!16$, $\hat{d}^{(l)}_{\mathcal{E}}\!=\!16$. We concatenate the representation $h^{(L)}_{\text{target\_route}}$, $h^{(L)}_{\text{candidate\_order}}$, and edge information between them as $h^{(L)}_{\text{consider}}$ (not applicable to GraphSage) and feed it into a fully-connected layer to predict the ranking score. CoMGNN and GNN-based baselines are trained according to softmax cross entropy loss \cite{bruch2019an}\footnote{The research work \cite{bruch2019an} provides proof that the softmax cross entropy loss is a bound on mean normalized discounted cumulative gain in log-scale.}. Besides, we compare CoMGNN with the state-of-the-art DNN-based methods for click-through rate (CTR) prediction task\footnote{Click-through rate prediction: a kind of items-of-interest recommendation task to predict whether the user will click-through an item.}: DIN \cite{DIN}, DIEN \cite{DIEN} and MIMN \cite{MIMN}. We feed the target route-order pair and historical route-order pairs to CTR methods. Each route-order pair includes attributes of the route, attributes of the order and calculated attributes between the route and the order. CTR methods are trained with the recommended settings in the published papers, where the objective function is the negative log-likelihood function.
We split the earlier $80$\% routes into training set, and evaluate the performance of comparison models on the rest $20$\% routes. We run each model $10$ times and report the average results of Recall@k and MAP (Mean Average Precision).

\begin{figure}
\centering
\includegraphics[width=\textwidth]{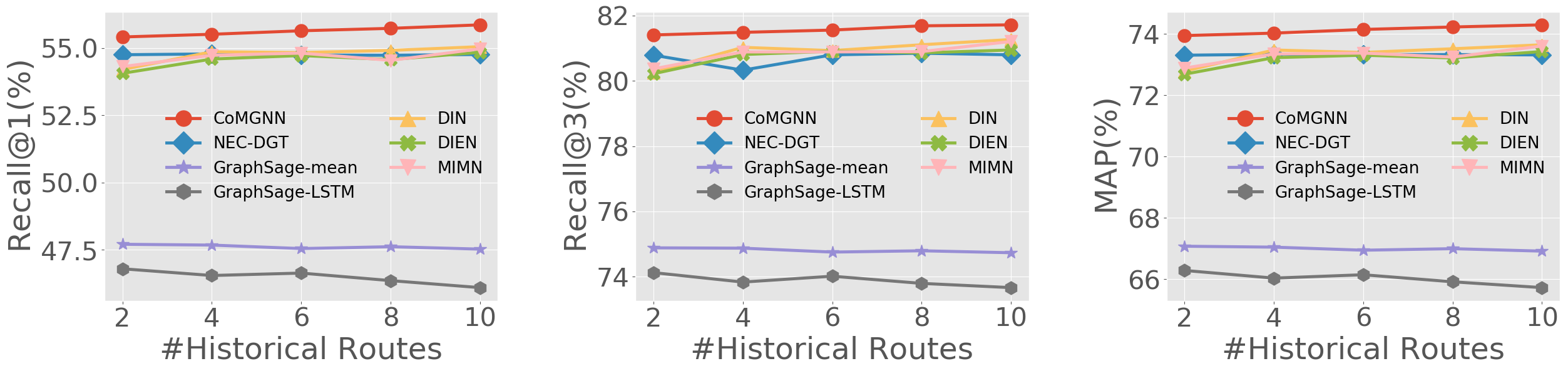}
\caption{Comparison on the hitch ride task with different numbers of historical routes.}
\label{fig:hitch_results}
\end{figure}

\paragraph{Results}
Figure \ref{fig:hitch_results} shows the test results with different numbers of latest historical routes $\mu$. In most of cases, the performance of methods increases with the increase of $\mu$. GraphSage methods (including GraphSage-mean and GraphSage-LSTM) perform worst, because they do not consider the edge attributes which are important for the hitch ride route matching task. When $\mu=2$, NEC-DGT performs better than other baselines, showing the advantage of graph structured information. Note that, DIN, DIEN and MIMN treat historical route-order pairs as sequential information and model them separately from the target route-order pairs. It makes the DNN-based models miss similarity information between the target route and historical routes. While DNN-based methods present competitive performance over NEC-DGT when $\mu>2$. Benefiting from the heterogeneous information and multi-attributions, our proposed CoMGNN outperforms all baselines under different $\mu$ settings, validating the efficacy of the proposed CoMGNN model.



\subsection{Traffic Prediction Task}
We build a multi-relational road graph for a city of eight million people, containing $74,685$ nodes. As shown in Figure~\ref{fig:hitch_ride}, each node on the graph representing a road segment, and edges between nodes are formed in three ways: 1) build a ``link to'' edge, when a road segment turns into another according to the commercial map ($124,489$  edges); 2) build a ``close to'' edge according to the physical distances between road segments ($363,833$ edges); 3) build a ``likely go to'' edge, when cars on a road segment are likely to flow to another within few steps according to historical trajectories ($206,339$ edges). In addition, we include the reverse relations of aforementioned edges.

From a ride-hailing platform, we collect the floating-car data from May 1st to July 31st, 2019. The mean speeds of passing vehicles on each road segment in every $5$ minutes are used as the spatiotemporal input on vertices. Then, dynamic features on edges are formed by averaging or calculating absolute difference between pairs of nodes they connect. In addition, static attributes on nodes (e.g., the width, length, free flow speed of road segment) and edges(e.g., turning direction, angles between two road segments) are collected to train the meta learners.

To measure the methods in different conditions, we construct two datasets UT\_JUNE and UT\_JULY, which contain data of $3$ weeks ranging from June 9th to June 29th, 2019 and July 9th to July 29th, 2019 respectively. For each dataset, we use the first week for training, the second week for validation and the last week for testing.


\paragraph{Experimental setup.}
We adopt Mean Absolute Percentage Errors (MAPE) as the target to minimize and form the loss function as $\mathcal{L} = \frac{|\hat{y} - y|}{y}+\lambda||\Theta||^2_2$.
where $y$ and $\hat{y}$ are the true and predicted traffic conditions respectively, and $\lambda$ denotes the rate for $L_2$ regularization, which is set to $10^{-5}$ in our experiments.
We compare our framework ST-CoMGNN with the following baselines: 1) GRU\cite{GRU}; 2) DCRNN\cite{DCRNN}; 3) STGCN\cite{STGCN}; 4) GWN$^*$\cite{GraphWaveNet}\footnote{Due to memory limits, we remove the self-adaptive adjacency matrix from original Graph WaveNet, as it is a dense matrix of size $|V|\times|V|$.}; 5) ASTGCN$^*$\cite{ASTGCN}\footnote{We revise the attention operation according to the implementation in GAT, reducing the space complexity from $O(|V|^2)$ to $O(|E|)$.}. Moreover, we perform ablation studies by removing heterogeneous information, edge features or meta graph attention module from ST-CoMGNN.    
To measure the performance of competing methods, we adopt Mean Absolute Percentage Errors (MAPE), Mean Absolute Errors (MAE) and Root Mean Squared Errors (RMSE) as evaluation metrics. During the training process of each method, we select the model with the best MAPE on validation set. In this task, we run each approach $3$ times and report the average results.

\paragraph{Results.}
Table~\ref{tab:speed_pred} lists the comparison of different methods for traffic condition forecasting in the next $5$, $15$ and $30$ minutes. As expected, all the spatiotemporal graph convolution based methods show better performances than GRU, which demonstrates the importance of structural information. ASTGCN$^*$ shows relatively better performance on $15$ and $30$ minutes ahead forecasting, which indicates that daily-periodic and weekly-periodic information is more useful for long-term prediction. 
Our proposed ST-CoMGNN and its variations outperform all the baseline methods by all the metrics, which strongly proves the effectiveness of our framework. From ablation studies, we demonstrate the importance of heterogeneous information, edge features and meta graph attention according to the results by the MAPE metric. And among them, heterogeneous information is relatively more beneficial. However, as we use MAPE as the optimization target, the difference among ST-CoMGNN and its variations by MAE and RMSE is not obvious.

\begin{table}[t]
\caption{Comparison Results on Traffic Speed Prediction}
\label{tab:speed_pred}
\centering
\resizebox{\textwidth}{!}{
\begin{tabular}{c|ccc|ccc|ccc}
\hline
Dataset&  \multicolumn{9}{c}{UT\_JUNE(5min/15min/30min)} \\                
Metric& \multicolumn{3}{c}{MAPE}&     \multicolumn{3}{c}{MAE}&      \multicolumn{3}{c}{RMSE}      \\
\hline                    
GRU&  35.14$\%$&  38.36$\%$&  40.58$\%$&  2.20&  2.36& 2.47& 3.05& 3.25& 3.38  \\
DCRNN&  33.68$\%$&  36.31$\%$&  38.30$\%$&  2.17& 2.31& 2.43& 3.01& 3.19& 3.34  \\
STGCN&  29.63$\%$&  31.99$\%$&  33.49$\%$&  1.94& 2.05& 2.13& 2.80&  2.97& 3.07  \\
GWN$^*$&  30.61$\%$&  33.06$\%$&  34.72$\%$&  1.97& 2.09& 2.20&  2.80&  2.96& 3.10 \\
ASTGCN$^*$& 29.78$\%$&  31.76$\%$&  32.64$\%$&  1.92& 2.04& 2.06& 2.75& 2.91& 2.94  \\
\hline                    
ST-CoMGNN&  \textbf{28.84$\%$}& \textbf{30.97$\%$}& \textbf{32.09$\%$}& \textbf{1.87}&  \textbf{1.99}&  \textbf{2.03}&  \textbf{2.71}&  \textbf{2.86}&  \textbf{2.93} \\
w/o het info& 29.22$\%$&  31.32$\%$&  32.26$\%$&  1.90&  \textbf{1.99}&  2.04& 2.74& 2.87& 2.94  \\
w/o edge info&  28.99$\%$&  31.17$\%$&  32.20$\%$&  1.89& \textbf{1.99}&  2.04& 2.72& 2.88& \textbf{2.93} \\
w/o meta att& 28.99$\%$&  31.04$\%$&  32.14$\%$&  1.88& 2.00&  \textbf{2.03}&  2.72& 2.88& \textbf{2.93}\\
\hline                    
\hline                    
Dataset&  \multicolumn{9}{c}{UT\_JULY(5min/15min/30min)} \\                
Metric& \multicolumn{3}{c}{MAPE}&     \multicolumn{3}{c}{MAE}&      \multicolumn{3}{c}{RMSE}      \\
\hline                    
GRU&  33.91$\%$&  37.06$\%$&  39.12$\%$&  2.11& 2.27& 2.35& 2.94& 3.14& 3.23  \\
DCRNN&  32.13$\%$&  34.71$\%$&  36.46$\%$&  2.06& 2.18& 2.29& 2.88& 3.04& 3.17  \\
STGCN&  28.69$\%$&  30.88$\%$&  32.30$\%$&  1.85& 1.94& 2.04& 2.69& 2.82& 2.96  \\
GWN$^*$&  29.69$\%$&  31.98$\%$&  33.55$\%$&  1.90&  2.01& 2.10&  2.69& 2.85& 2.97  \\
ASTGCN$^*$& 28.85$\%$&  30.65$\%$&  31.57$\%$&  1.84& 1.92& 1.98& 2.64& 2.76& 2.82  \\
\hline
ST-CoMGNN&  \textbf{27.97$\%$}& \textbf{29.92$\%$}& \textbf{30.86$\%$}& \textbf{1.80}&  1.90&  \textbf{1.93}&  \textbf{2.61}&  2.75& 2.80 \\
w/o het info& 28.20$\%$&  30.10$\%$&  31.01$\%$&  1.81& \textbf{1.89}&  1.95& 2.62& \textbf{2.74}&  2.81  \\
w/o edge info&  28.03$\%$&  29.96$\%$&  30.92$\%$&  1.81& 1.90&  \textbf{1.93}&  \textbf{2.61}&  \textbf{2.74}&  \textbf{2.79} \\
w/o meta att& 28.07$\%$&  29.97$\%$&  30.93$\%$&  \textbf{1.80}&  \textbf{1.89}&  1.94& \textbf{2.61}&  \textbf{2.74}&  2.81  \\
\hline                    
\end{tabular}}
\end{table}

\section{Conclusion}
In this paper, we introduce a novel deep learning framework CoMGNN on the heterogeneous graph, where node and edge states are co-evolved. Moreover, we propose an enhanced version called ST-CoMGNN for spatiotemporal prediction. Experiments show that our models outperform other state-of-the-art methods on two real-world tasks, including hitch ride route matching and traffic condition prediction. In the future, we will apply CoMGNN to other tasks and construct more powerful multi-attributed graphs, such as adding collaborative information to the graph for hitch ride route matching task.



\newpage
\section*{Broader Impact} 
This work has the following potential positive impacts. 1) Existing GNN applications with intricate information will possibly benefit from the proposed framework, e.g., recommendation systems, traffic information systems, etc. 2) This work may inspire further researches to focus on how to effectively take advantage of various information.

At the same time, this work may have some negative consequences. 1) Introducing additional information is not always helpful. Sometimes, the noise in data could decrease the performance of systems. Moreover, processing additional information requires more computing resources or longer computing time. 2) This work points out how to encode more information, but the abuse of information may lead to personal privacy problems. Users need to ensure that all the input privacy data is authorized in advance.


\bibliographystyle{plain}
\bibliography{paper}

\end{document}